\documentclass[journal]{IEEEtranTIE}
\usepackage{graphicx}
\usepackage{cite}
\usepackage{picinpar}
\usepackage{amsmath}
\usepackage{url}
\usepackage{flushend}
\usepackage[latin1]{inputenc}
\usepackage{colortbl}
\usepackage{soul}
\usepackage{multirow}
\usepackage{pifont}
\usepackage{color}
\usepackage{alltt}
\usepackage[hidelinks]{hyperref}
\usepackage{enumerate}
\usepackage{siunitx}
\usepackage{epstopdf}
\usepackage{pbox}
\usepackage{bm}
\usepackage{amsthm,amssymb,mathrsfs,mathtools,amsmath}
\usepackage{algorithmicx, algpseudocode}  
\usepackage{algorithm}    
\usepackage{setspace}
\begin{document}
\title{A Versatile Pseudo-Rigid Body Modeling Method}

\author{
	
	Amir Molaei,
	Amir G. Aghdam, and
	Javad Dargahi

	\thanks{A. Molaei and J. Dargahi are with the Department
		of Mechanical, Industrial and Aerospace Engineering, Concordia University, Montreal, Canada, e-mail: {\tt\small a\_molaei@encs.concordia.ca}, {\tt\small dargahi@encs.concordia.ca}}
	\thanks{A. G. Aghdam is with the Department
		of Electrical and Computer Engineering, Concordia University, Montreal,
		Canada, e-mail: {\tt\small amir.aghdam@concordia.ca}}	

}
\maketitle
\begin{abstract}

A novel semi-analytical method is proposed to develop the pseudo-rigid-body~(PRB) model of robots made of highly flexible members (HFM), such as flexures and continuum robots, with no limit on the degrees of freedom of the PRB model. The proposed method has a simple formulation yet high precision. Furthermore, it can describe HFMs with variable curvature and stiffness along their length. The method offers a semi-analytical solution for the highly coupled nonlinear constrained optimization problem of PRB modeling and can be extended to variable-length robots comprised of HFM,  such as catheter and concentric tube robots. We also show that this method can obtain a PRB model of uniformly stiff HFMs, with only three parameters. The versatility of the method is investigated in various applications of HFM in continuum robots. Simulations demonstrate substantial improvement in the precision of the PRB model in general and a reduction in the complexity of the formulation.

\end{abstract}\
\begin{IEEEkeywords}
Continuum robots; pseudo-rigid body model; flexible robots; concentric tube robot; steerable catheter; compliant mechanism; highly flexible elements 
\end{IEEEkeywords}
{}    	
\IEEEpeerreviewmaketitle
	
\section{Introduction}
\label{Intro}

Pseudo rigid body (PRB) theory is a modeling approach that provides a rigid body equivalent of highly flexible members (HFM) with large deflection under various loading conditions. Using PRB modeling method, a HFM can be replaced with $n$ hinged rigid segments with virtual springs at the joints, and $n$ in fact, is the degrees-of-freedom (DoF) of the PRB model. Thus, the force-deflection behavior of the HFM can be obtained using algebraic equations through the PRB model, which is primarily a boundary value problem (BVP) in the continuum representation. From the mathematical perspective, a PRB equivalent model of a continuum element can be formulated as an optimization problem, where the optimal values for the segment lengths and the stiffness of the virtual springs are to be found. This is done by formulating the static force mapping of the PRB model using its Jacobian matrix. This formulation, however, requires the deformation of the virtual springs. Thus, the tip deflection under a given tip load, which is found using the continuum model, can be used to formulate the inverse kinematics (IK) for finding each virtual spring deformation. This implies that PRB model is formulated as a two-objective optimization problem, which should satisfy the static force mapping over a range of loads, and at the same time, should meet the IK. By increasing the DoF of the PRB model, the problem of finding optimal parameters of the PRB model becomes more complex, as there is no unique analytical formulation for the IK of the serially hinged rigid segments with a DoF greater than three.  

The PRB theory was first introduced by Larry L. Howell in 1994 as an approximation technique for modeling a compliant/flexure mechanism \cite{howell1996evaluation}, and has been used for the modeling of robotics systems made of HFM in various applications. A PRB equivalent model allows the use of well-developed methods in rigid robotics to be applied to robots with flexible elements. Applications of this modeling approach can be found in a robotic fish, where the optimal compliance of the fish fin is determined for maximizing the thrust\cite{park2012kinematic}. In \cite{sanan2009robots}, PRB modeling method is used for a robot with inflatable links.
In another study \cite{doyle2011avian}, this method is used for the analysis and design of an avian-inspired passive perching mechanism for a robotic rotorcraft. This method has also been used for the modeling of new insertable robotic end-effectors platform for single port surgery \cite{ding2012design}.
The simplicity of the PRB model, compared to the continuum model, makes it an efficient method for the modeling of continuum robots. That is the primary reason why it has recently been used by several researchers in this area. The authors in \cite{satheeshbabu2017designing} use the method for the modeling of fiber-reinforced elastomeric enclosures that are fundamental building blocks of pneumatic soft robots. The work in \cite{khoshnam2013pseudo} employs a 3-DoF PRB model for the modeling of catheter tip force for the control purpose. In another work, the authors in \cite{venkiteswaran2019shape}, use this method to study the deformed shape and reaction forces of continuum manipulators interacting with their environment, and verified it experimentally. In \cite{huang20193d}, the 3D static modeling method and experimental verification of
continuum robots based on PRB theory have been investigated. The authors in \cite{greigarn2015pseudo} use this method to investigate the kinematics of MRI-compatible,
magnetically actuated, steerable catheters, and the experimental results are provided in \cite{greigarn2017experimental}. The work \cite{greigarn2015pseudo} also introduces a method of finding a
set of parameters for the PRB model from a set of experimental
data for the catheter. The authors in \cite{greigarn2018jacobian} use the PRB model of an MRI-actuated continuum robot for the quasi-static Jacobian-based task space motion planning. The PRB model has also been used in a recent study in needle steering for adaptive energy shaping control \cite{franco2021adaptive}.

The PRB modeling method was initially developed for HFMs subjected merely to end moment~\cite{howell1994method} or to end forces \cite{howell1995parametric}, assuming the tip of the HFM follows a circular path. Parametric approximation of the straight HFM tip deflection using PRB modeling method started with a 1-DoF model consisting of two rigid links hinged with a torsional spring, which had relatively large error values for large tip deflection angles \cite{howell1995parametric}. The accuracy of the PRB modeling method  was improved in \cite{su2009pseudorigid}, using a 3-DoF model with a maximum tip deflection error of $1.2\%$ compared to the FEA model. The authors in \cite{chen2011finding}, formulated the problem in the particle swarm optimization context with the methodology introduced in \cite{su2009pseudorigid} to increase the precision. The PRB modeling method has also been utilized to analyze the circular HFM \cite{venkiteswaran2016pseudo,venkiteswaran2018versatile}. In \cite{venkiteswaran2016pseudo}, a 2-DoF PRB model is proposed for the circular HFM with an error not exceeding 3.05\% in the tip position. The method is then extended to 3-DoF PRB model in \cite{venkiteswaran2018versatile} with a symmetrical kinematic structure and compliance about the central joint for a circular HFM with a uniform cross-section.

As the existing PRB modeling methods are formulated for specific case studies, the limitations of these methods cannot be clearly identified. To have a better insight into the shortcoming of the PRB modeling in literature, we have generalized a recent PRB modeling formulation which is introduced in \cite{venkiteswaran2018versatile}. This method offers a 3-DoF PRB model for constant curvature, uniformly-stiff HFM, assuming symmetrical length and stiffness around the central joint, for simplification of the problem. Our extended formulation (see Algorithm~\ref{alg:1}), considers an $n$-DoF PRB model for arbitrary-curved HFM with non-uniform stiffness along the length. Considering Algorithm~\ref{alg:1}, the limitations of the existing methods are as follows. 

It requires the IK solution for the PRB model, which does not have a unique analytical solution for a DoF more than three. Consequently, most of the existing methods for the PRB modeling are developed for three DoF, which limits the precision of the PRB model. The higher the DoF, the higher the precision of the model \cite{su2009pseudorigid}. Additionally, the initial position of the joints in the no-load condition should be specified, which is supposed to satisfy the equation representing the centerline of HFM. This assumption will be formulated as multiple equality constraints in the optimization problem. Moreover, the existing formulation requires additional constraints on the positiveness of the stiffness and length of the elements with proper upper bounds. More importantly, the resultant optimization problem involves, in general, a two-objective, highly nonlinear equation that is to be solved numerically. 
One other shortcoming of the existing methods in the literature is that there is no direct method for the PRB modeling of variable curvature HFMs to the best of the authors' knowledge. Such modeling is generally done by dividing the variable curvature into circular segments and then finding the PRB model for each circular segment. One reason behind this limitation is the complexity of the initialization of the joint position in the no-load condition. For HFMs with circular shape, the length of each rigid segment of the PRB model is equal to the cord of a circle which is a fraction of the total angle of the circular HFM. However, for an arbitrary curve, the length of the segment cannot simply be determined, which leads to a complex set of nonlinear equations. Furthermore, there is no unified, comprehensive method that can be used for the general PRB modeling problem, and most of the studies are limited to specific case studies such as straight HFMs. Moreover, there are applications, such as concentric tube robots (CTR), in which the length of the HFM changes. The available methods fail to cover such applications suitably.

To overcome the shortcomings of the existing literature discussed, we propose a novel methodology for the PRB modeling, which can easily be used in various applications. In the proposed method, the BVP corresponding to the continuum model, which is used only to find the tip deflection in the existing methods, is used to solve the IK. As a result the two cost functions are decoupled, and the optimization problem in PRB modeling becomes a single objective one, specified by the static force mapping equation. Then we propose an analytical solution to the optimization problem, which provides the optimal values of stiffness of the virtual springs for an $n$-DoF PRB model. Through different case studies, we show how the proposed method can be used for continuum robots with variable curvature and stiffness along the length, for both fixed and variable lengths. The results of our studies also provide an $n$-DoF PRB model for arbitrarily-curved uniformly-stiff HFM with just three parameters. 

The remainder of the paper is structured as follows. In Section~\ref{euler}, the Euler beam theory for the general case of HFM is formulated. In Section~\ref{PRB}, we provide a general formulation for the PRB modeling. Then in Section \ref{proposed}, the proposed methodology is described in detail. Next in Section \ref{result}, the application of the proposed method is investigated for five different case studies, considering straight, circular, arbitrarily curved, and non-uniformly-stiff, as well as variable-length HFMs. Finally, the concluding remarks are given in Section~\ref{conclusion}.

\section{Force/deflection Modeling of HFMs}
\label{euler}

\begin{figure}[ht]
	\centering
	\includegraphics[height=4cm,trim={1cm 6.5cm 1cm 5.7cm},clip=true]{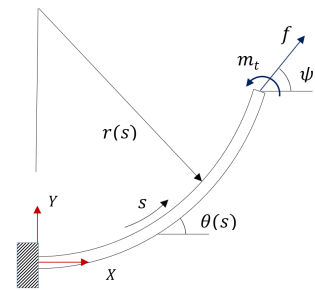}
	\caption{An initially curved HFM with a load at the tip}
	\label{CurvedBeam}
\end{figure}

An initially curved HFM with the total length of $S$ under the applied force $f$ and moment $m_t$ is demonstrated in Fig.~\ref{CurvedBeam}. The HFM has a variable curvature $r(s)$ and stiffness along the length. The stiffness of the HFM, which is referred to as flexural rigidity, is a function obtained by multiplication of the elasticity module $E$ of the material and the second moment of area $I$; either of them could change along the length. For the sake of generality, we represent variable flexural rigidity as a function of length defined by $EI(s)$ and fixed flexural rigidity by $EI$.
As the deformation of the HFM is mainly due to the shear moment, the Euler beam theory suitably describes its behavior. With an applied load at the tip, the bending moment $m(s)$ along the deformed HFM is given by: 
\begin{equation}
\begin{cases}
\label{eq:1111}
m(s)= m_{t}+f(x_{t}-x)\sin{\psi}-f(y_{t}-y)\cos{\psi}\\
x(s)=\int_{0}^{s}\cos{\theta(\lambda)}d\lambda\\
y(s)=\int_{0}^{s}\sin{\theta(\lambda)}d\lambda
\end{cases}
\end{equation}
where $\bm{p}=[x_t,y_t,\theta_t]$ represents the tip position vector of the continuum model, and $\lambda$ is the integral variable.
The shape of the HFM under the applied load at the tip can be expressed using the Euler beam theory as:
\begin{equation}
\label{eq:2}
{\theta^{\prime}}(s)= \frac{m(s)}{EI(s)}+\frac{1}{r(s)}
\end{equation}
Using Eq.~\eqref{eq:1111} and differentiating~\eqref{eq:2} with respect to $s$ results in:
\begin{equation}
\begin{cases}

{\theta^{\prime\prime}}(s)= \frac{m^\prime(s)}{EI(s)}-\frac{EI^\prime(s)m(s)}{EI^2(s)}-\frac{r^{\prime}(s)}{r^{2}(s)}\\

\theta(s=0)=0\\
{\theta^{\prime}}(s=S)=\frac{m(S)}{EI(S)}+\frac{1}{r(S)}
\end{cases}
\label{eq:3_}
\end{equation}

Equation~\eqref{eq:3_} is the general governing equation of a 2D HFM under tip load. This can be used for the modeling of a HFM which has variable curvature and variable stiffness along the length. When dealing with constant curvature $R(s)=R$ and uniform stiffness, Eq.~\eqref{eq:3_} is simplified as:
\begin{equation}
\begin{cases}

{{\theta^{\prime\prime}}(s)=\frac{f}{EI}\sin{(\theta-\psi)}}\\
\theta(s=0)=0\\
{\theta^{\prime}}(s=S)=\frac{m(S)}{E I}+\frac{1}{R}
\end{cases}
\label{eq:4}
\end{equation}

Note that Eq.~\eqref{eq:3_} is computationally complex, in general. This is important for real-time control of flexible robots interacting with a soft environment. Additionally, force control schemes require the knowledge of the Cartesian stiffness of the robot \cite{yasin2021joint,yip2016model,mahvash2011stiffness}, which cannot simply be obtained using~\eqref{eq:3_}. By using a PRB model, the formulation of the Cartesian stiffness of the robot is simplified. Additionally, the stiffness model can be used to estimate the interaction force via deflection measurement.

\section{PRB Modeling of HFMs}
\label{PRB}
\begin{figure}[ht]
	\centering
	\includegraphics[height=4cm,trim={7cm 14.5cm 7cm 8.1cm},clip=true]{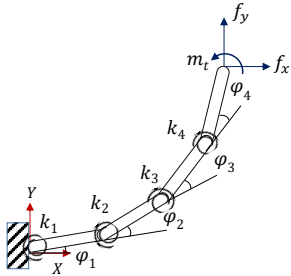}
	\caption{A 4-DoF PRB model of a curved HFM}
	\label{RigidBodyModel}
\end{figure}

For the PRB modeling of a curved HFM with a total length of $S$, it is first divided into a finite number of rigid segments. An $n$-DoF PRB model includes $n$ rigid segments connected to each other via $n$ revolute joints with torsional springs, in general. The stiffness of each revolute joint is represented by $k_i$, and the length of each segment is $l_i$ for $i=1:n$. We use $\bm K_{\varphi}$ as the diagonal matrix of the joint stiffness $diag(k_1,...,k_n)$ and $\bm l$ as the vector of the segments' length, i.e. $[l_1,..,l_n]^T$. Fig.~\ref{RigidBodyModel} shows a 4-DoF PRB model, which is connected to a fixed base via a torsional joint.

In this figure, the relative angle of each segment with respect to its adjacent segment is denoted by $\varphi_i$, with ${\bm{\varphi}}$ being the vector of joint variables, i.e. , $[\varphi_1,...,\varphi_n]^T$. Note that $\varphi_1$ is measured with respect to the horizontal line. Introducing $\hat\theta_j=\sum_{i=1}^{j}{\varphi_i}$ for $i=1:n+1$ as the angle of each segment with respect to the horizontal line, the tip position of each segment with respect to the fixed Cartesian frame, depicted in Fig.~\ref{RigidBodyModel} can be written as:
\begin{equation}
\begin{cases}
\hat x_i=\sum_{j=1}^{i}{l_j \cos \hat \theta_j}\\
\hat y_i=\sum_{j=1}^{i}{l_j \sin \hat \theta_j}\\
\end{cases}
\label{eq:segment_tip}
\end{equation}
 It is notable that we use$~\hat~$ to differentiate between the variables in the continuum and PRB model. Moreover, as there is no joint at the tip, we refer to the tip angle of the PRB model by $\hat \theta_{n+1}$ by considering a virtual segment with a length of zero. Using the above formulation, the  tip position vector $\hat{\bm{p}}$ of the PRB model, i.e. $[{\hat x}_{n},{\hat y}_{n},{\hat \theta}_{n+1}]^T$, is expressed by: 
\begin{equation}
\begin{cases}
\label{InverseKinematic}
{\hat x}_{n}=\sum_{j=1}^{n}{l_j \cos \hat \theta_j}\\
{\hat y}_{n}=\sum_{j=1}^{n}{l_j \sin \hat \theta_j}\\
\hat \theta_{n+1} =\sum_{i=1}^{n+1}{\varphi_i}
\end{cases}
\end{equation}
with $\varphi_{n+1}$ being a fixed offset value, which is the difference of the tip angle of the continuum model and the PRB model in the no-load condition.

Let the applied wrench at the tip ($\bm{w}=[f_x ,f_y,m_t]^T$) be mapped to the torsional force of the virtual springs ($\bm{\tau}=[{\tau}_1,...,{\tau}_n]^T$) at the joints using the Jacobian matrix ($\bm J$) as: 
\begin{equation}
    {\bm{\tau}} = \bm{J}^{T}\bm{w}
    \label{Jacobian}
\end{equation}
Each element of $\bm{\tau}$ in the above equation is proportional to the deflection of the corresponding joint $\delta\varphi_i$ under the applied load, i.e., $\tau_i=k_{i}\delta\varphi_i$. This can be written in the matrix form as $\bm{\tau} = \bm{K}_{\varphi}\bm{\Delta}\bm{\varphi}$, with $\bm{\Delta}\bm{\varphi}= [\delta\varphi_1, ...,\delta\varphi_n]^T$. On the other hand, the Jacobian matrix is by definition, expressed as $\bm{J}={\partial \bm{\hat{p}}}/{\partial \bm\varphi }$.

\subsection{Parameter Optimization of PRB Model} 
\label{literaturePRB}

   It is now desired to derive a PRB model analogous to that of a continuum model with the highest accuracy in describing the tip point displacement and the compliance behavior under a wide range of loads. This constitutes an optimization problem which  can be expressed by two sets of cost functions, $E_f= \frac{1}{N}\sum_{q=1}^{N} \lVert{\bm{K_{\varphi}{\Delta\varphi}}-\bm{J^T w}}\lVert^q$ and $E_x= \frac{1}{N}\sum_{q=1}^{N} \lVert{\bm{p}-\bm{\hat{p}}}\lVert^q$, for the force and position errors; in which $q$ represents each loading condition with a specific wrench ($[f_x,f_y,m_t]^q$) and $N$ is the total number of the loading conditions. We represent the no-load condition $\bm {w=0}$ with $q=0$. In this optimization problem $\bm{\Delta{\varphi}^q}$ is the optimization variable and $\bm l$ and $\bm K$ are the unknown parameters. 
The $\bm{\Delta{\varphi}^q}=\bm{\varphi}^q-\bm{\varphi}^0$, requires $\bm{\varphi}^q$ and $\bm{\varphi}^0$, which are obtained through the IK solution of the PRB model in each loading condition and its no-load configuration, respectively.
On the other hand, $\hat{\bm{p}}$ and $\bm{J^T}$ are defined as functions of $\bm{\varphi}^q$, $\bm{\varphi}^0$ and $\bm l$, i.e. $\hat{\bm{p}}=\hat{\bm{p}}{(\bm{\varphi}^q,\bm{\varphi}^0,\bm l)}$ and $\bm{J^T} =\bm{J^T}{(\bm{\varphi}^q,\bm{\varphi}^0,\bm l)}$. Generally, for the curved HFMs as in \cite{venkiteswaran2018versatile} it is assumed that in the no-load condition, the joints of the PRB model lie on the centerline of the continuum model curve. The centerline of the curve is a continious function of $x$ and $y$ which can be defined as $g(x,y)=0$. This assumption works as an additional constraint, such that the relation of $g(\hat x_i^0,\hat y_i^0) = 0,~i=1:n$ holds in the optimization procedure, in which $(\hat x_i^0,\hat y_i^0)$ represents the no-load position for each of the PRB model joints. The equation representing function $g(x,y)$ could be complex, in general
Note that the length of the segments $l_i$ and stiffness of the joints $k_i$ cannot have negative values for physical systems. Thus, the optimization problem for the PRB modeling of HFMs with variable curvature can be formulated as follows:

\begin{equation}
\begin{cases}
\label{Eq:CostFunctionMain}
E_f= \frac{1}{N}\sum_{q=1}^{N} \lVert{\bm{K{\Delta\varphi}}-\bm{J^T w}}\lVert^q\\ 
E_x= \frac{1}{N}\sum_{q=1}^{N} \lVert \bm{p}-\bm{\hat{p}}\lVert^q\\
k_i>0,~l_i>0\\

\begin{cases}
\bm{\hat{p}}^q= [\hat x_n^q,\hat y_n^q,\hat \theta_{n+1}^q]^T\\
\bm{{p}}^q= [x_t^q,y_t^q,\theta_t^q]^T\\

\hat x_n^q=\sum_{j=1}^{n}{l_j \cos\hat \theta_j^q}\\
\hat y_n^q=\sum_{j=1}^{n}{l_j \sin\hat \theta_j^q}\\
\hat \theta_j^q=\sum_{i=1}^{j}{\varphi_i^q}
\end{cases}\\

\begin{cases}
g(\hat x_i^0,\hat y_i^0) = 0\\
\hat x_i^0=\sum_{j=1}^{i}{l_j \cos\hat \theta_j^0}\\
\hat y_i^0=\sum_{j=1}^{i}{l_j \sin\hat \theta_j^0}\\
\hat \theta_j^0=\sum_{i=1}^{j}{\varphi_i^0}
\end{cases}
\end{cases}
\end{equation}

Due to the strong coupling of the cost functions $E_f$ and $E_x$ to the IK, for the sake of simplicity, most of the existing results in the literature consider 3-DoF PRB models. Note that for an $n$-Dof PRB model with $N$ loading conditions, $n(N+2)$ unknowns are to be found, e.g., more than 3 million unknowns for a 100-DoF PRB model over 3000 loading conditions. As discussed, the optimization problem, defined by Eq.~\eqref{Eq:CostFunctionMain} also requires the IK for the no-load condition, and due to its complexity, merely the circular or straight HFMs are studied in the literature such as \cite{venkiteswaran2018versatile}. In the above work, it is also assumed that the first and the third joint have the same stiffness and length, i.e., $k_1=k_3$ and $l_1=l_3$, which limits the search algorithm.

From the above discussion and Eq.~\eqref{Eq:CostFunctionMain} finding an arbitrary $n$-DoF PRB model, in general, is a complex problem. Therefore the PRB modeling methods in the literature are limited to fixed-curvature HFMs with uniform stiffness distribution along the length with limited DoF. Furthermore, most of the existing methods have moderate accuracy, which may not be acceptable in many medical applications. Algorithm \ref{alg:1} describes the formulation discussed above for the PRB modeling. To overcome the shortcomings, we propose a different yet straightforward semi-analytical approach in the next Section

\begin{algorithm}
\small{
    \caption{General method for the PRB modeling of HFMs}
    \label{alg:1}
    \begin{algorithmic} [1]
    
        \State {initialize $S>0$, $EI(s)$, $r(s)$, and $n\in \mathbb{N}$}
        \State {$g(x,y)$ $\leftarrow$ use $r(s)$}
        \Require {$w_q$ for $q = 1:N$,} \Comment{$w_q$s should cover the actual loading conditions of the HFM}
        \State {symbolically formulate $\hat{\bm p}$ as a function of $\bm l$ and $\bm \varphi$ using \eqref{eq:segment_tip}}
        \State {symbolically formulate $\bm J$ for the $n$-DoF PRB model}
        \State {symbolically fromulate $\bm \varphi^0$ as a function of $\bm l$ using $g$}
        \For{$q = 1 : N$}
            \State {$\bm p$ $\leftarrow$ solve~\eqref{eq:3_}~for~$\bm w_q$}
            \State {$E_f^q=\lVert \bm{p}-\bm{\hat{p}}\lVert^q$}
            \State {$E_p^q=\lVert{\bm{K{\Delta\varphi}}-\bm{J^T w}}\lVert^q$}
        \EndFor
        \State {$E_p= \frac{1}{N}\sum_{q=1}^{N} E_f^q$} 
        \State {$E_f= \frac{1}{N}\sum_{q=1}^{N} E_f^q$} 
        \State{Set the constraints as: $k_i>0$ \& $l_i>0$ \& $\sum_{i=1}^{n}l_i\leq s$} 
        \State {Minimize $E_p$ \& $E_f$ and find the optimal $\theta_i^q$s, $k_i$s and $l_i$s }        
    \end{algorithmic}}
\end{algorithm}

\section{A Versatile PRB Modeling Framework}
\label{proposed}
\begin{figure}[ht]
	\centering
	\includegraphics[height=5cm,trim={3cm 8cm 3cm 8cm},clip=true]{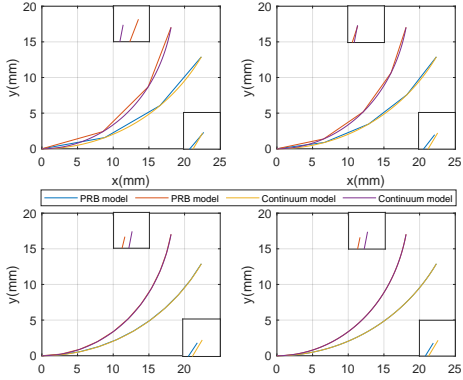}
	\caption{Deformation of the continuum model compared to PRB models with 3, 4, 5 and 30-DoF}
	\label{tipdisplacement}
\end{figure}

The main idea behind the proposed approach herein is to decouple the IK from the optimization procedure. Thus, the values of $l_1, ..., l_n$ are first found, and subsequently, the optimal stiffness $k_1, ..., k_n$ associated with them are obtained analytically using $E_f$. To this aim, we use Eq.~\eqref{eq:3} for solving the IK, which describes the mechanical behavior of HFMs under tip loading. This approach results in realistic values for optimization variables compared to the existing methods, which formulate IK with a geometric equation without capturing the physical behavior of HFMs. Additionally, in our method, as the physics is included in the IK formulation, the behavior of the PRB model along the length is also close to the realistic behavior. It is in contrast with the existing methods, where the PRB model is obtained just by matching the displacement of the tip point with that of the continuum model, and deflection of the HFM along the length is not taken into account. In the proposed approach, first, the DoF of the PRB model $n$ is chosen. It is to be noted that there is no limit on the maximum value of $n$ in the proposed method. The larger the $n$, the higher the precision of the method. Secondly, we divide the length of the HFM into $n$ segments $[s_1,...,s_n]$, which results in $n+1$ points along the curve length. Then, the governing continuum equation of the HFM is numerically solved for the given load, while the curve is divided into $n$ segments, where the length of segment $i=1:n$ is $s_i$. As the first step of the proposed method we let $s_i = l_i$. In other words, the length of each segment of the PRB method is specified, which is the first step for solving the IK. Solving the inverse kinematic for the PRB model also requires the angle of each segment. To this end, we introduce a new representation of the deformation of the HFM, which is represented in the length coordinate by Eq.~\eqref{eq:3_} in Cartesian coordinates. This is done by incorporating $x$ and $y$ with the corresponding boundary conditions in Eq.~\eqref{eq:3_}, which results in Eq.~\eqref{eq:3}. 

\begin{equation}
\begin{cases}
\label{eq:3}
{\theta^{\prime\prime}}(s)= \dfrac{m^\prime(s)}{EI(s)}-\dfrac{EI^\prime(s)m(s)}{EI^2(s)}-\dfrac{r^{\prime}(s)}{r^{2}(s)}\\
x^\prime(s) = cos\theta(s)\\
y^\prime(s) = sin\theta(s)\\
\theta(s=0)=x(s=0)=y(s=0)=0\\
{\theta^{\prime}}(s=l)=\dfrac{m(l)}{EI(l)}+\dfrac{1}{r(l)}\\
\end{cases}
\end{equation}
By solving the above equation with BVP solvers for the $n+1$ points along the length, one can find the corresponding $(\hat x_{i},\hat y_{i})$. Thus, the absolute angle of each segment $\hat \theta_i$ is found using the following equality:
\begin{equation}
 \theta_i=\arctan(\frac{\hat y_i-\hat y_{i-1}}{\hat x_i-\hat x_{i-1}}),~i=1, ..., n
 \label{segment_angle}
\end{equation}
Using Eq.~\eqref{eq:3} for the IK ensures that the absolute value of the position error ($E_x$) is minimized, and at the same time, $E_f$ is decoupled from $E_x$. Thus the optimization of the PRB model requires minimizing $E_f$. 
For a better insight into the working principle of the proposed method for minimizing $E_x$, we consider a pre-curved HFM under two different loading conditions, resulting in large deflection. The HFM used in this simulation study is the inner tube of a CTR, with the parameters specified in Table~\ref{param}. The simulation is done for two different loading conditions, $10~mN$ and $100~mN$ in magnitude, both applied at the tip with an angle of $156^\circ$ with respect to the horizontal plane. For the given loads, we have compared the shape of the continuum model and the PRB models with 3, 4, 10, and 30-DoF in Fig.~\ref{tipdisplacement}. For better visualization of the tip position error, the tip point is confined in a square with a side length of $0.0008~mm$ enlarged 6250 times. 
Considering the length of the enlarged square, it is apparent that the position error is less than $0.00001~mm$ for all the cases. For the angle error, it is notable that this requires adding an offset angle to the PRB model based on the tip angle difference in the no-load condition.

In the following we investigate the analytical solution of the $E_f$ which can expanded as the following form:
\begin{equation}
\label{Eq:CostFunctionfor3}
\sum_{q=1}^{N} ((k_1{\Delta\varphi_1}-\bm{J}_{1}^T\bm{w})^{2} + ...+ (k_n{\Delta\varphi_n}-\bm{J}_{n}^T\bm{w})^{2})_q
\end{equation}

where $\bm J_{i}^T$ is the $i$th row of the $\bm J^T$. As it is clear from the equation all the $k_i$ are decoupled and $E_f= \sum_{i=1}^{n}E_{f_i}$, in which $E_{f_i}$ is:
\begin{equation}
\label{Eq:CostFunction_3}
E_{f_i}= \sum_{i=1}^{N} (k_i{\Delta\varphi_i}-{\bm{J}}^T_{i}\bm{w})^{2}_q
\end{equation}
Considering the above equations, the minimum value of $E_f$ can be found using:
\begin{equation}
\frac{\partial{E_f}}{\partial{\bm{k}}}=[\frac{\partial{E_{f_1}}}{\partial{{k}_1}},...,\frac{\partial{E_{f_n}}}{\partial{{k}_n}}]^T=0
\end{equation}
Thus the optimal value of $k_i$ will be as follows:
\begin{equation}
\label{optimalstiffness}
k_i=\frac{\sum_{p=1}^{N}({\bm{J}}^T_{i}\bm{w}{\Delta\varphi_i})_p}{\sum_{p=1}^{N} {({\Delta\varphi_i}}^2)_p}
\end{equation}
As mentioned earlier for finding the minimum value of $E_f$, the minimum value of each $E_{f_i}$ should be found. Expanding $E_{f_i}$ results into a quadratic function of $k_i$ as follows:
\begin{equation}
\label{Eq:Quadratic}
{k}^2_i\sum_{q=1}^{N}({\Delta\varphi_i})^2_q-2{k_j}\sum_{q=1}^{N}({\bm{J}}^T_{i}\bm{w}{\Delta\varphi}_i)_q +\sum_{q=1}^{N}({\bm{J}}^T_{i}\bm{w})^{2}_q
\end{equation}

As we discuss through the following examples, since there is no limit on the DoF in our proposed methodology, it is more convenient to consider equal length for the segments and increase the precision of the PRB model by increasing DoF. There are also other advantages for the PRB model with equal length of the segments, which will be discussed in the examples. For the equal length of the segment~Eq.~\eqref{optimalstiffness} explicitly gives the optimal values of the stiffness for the PRB model. Thus, the PRB model optimization simplifies to a great extent, which at the same time offers higher precision, and can work for the various case of flexible members problems. The proposed method is summarized as Algorithm \ref{alg:2}. In the next section, we show the application of the proposed method for different case studies in surgical robotics.

\begin{algorithm}
\small
    \caption{Proposed method for the PRB modeling of HFMs}
    \label{alg:2}
    \begin{algorithmic} [1]
    
        \State {initialize $S>0$, $EI(s)$, $r(s)$ and $n\in \mathbb{N}$}
        \Require {$w_q$ for $q \in \mathbb{N}$} \Comment{$w_q$s should cover the actual loading conditions of the HFM}
        \State{$l_i=l$ and $l=\frac{S}{n}$ }
        \State {$\hat x_i^0$ \& $\hat y_i^0$ ~for~$i=1:n$ $\leftarrow$ solve~\eqref{eq:3_}~for~$w=[0,0,0]^T$}\
        
        \State{$\hat\theta_i^0$ ~for~$i\in \{1:n\}$ $\leftarrow$ solve~\eqref{segment_angle}}
        
        \For{$q = 1 : N$}
        \State { $\hat x_i$ \& $\hat y_i$ for~$i\in\{1:n\} \leftarrow$ solve~\eqref{eq:3_}~for~$w_q$~over $n+1$~equidistance points as $[0,l,...,nl]$ }
        \State{ $\hat{\theta}_i^q$ for~$i\in\{1:n\}$  $\leftarrow$ solve \eqref{segment_angle}}
        \State{$\delta{\varphi}_i^q=\hat{\theta}_i^q-\hat{\theta}_i^0$ for~$i\in\{1:n\}$}
        \State {$J^q$}
        \EndFor
        \State { $k_i$ for~$i\in\{1:n\}$  $\leftarrow$ solve \eqref{optimalstiffness}}
    \end{algorithmic}
\end{algorithm}

\section{Versatility of the Proposed PRB Modeling Framework }
\label{result}
To investigate the versatility of the proposed strategy for the PRB modeling, we consider five practical case studies. For the first case, we use the proposed method for the modeling of the catheter/guide-wire. A catheter is a thin wire inserted into the blood vessel for endovascular interventions, such as stenosis treatment. In catheterization The catheter interacts with the blood vessels all through the contact point with the arterial wall. Such interaction with the vessel might result in large contact forces, damaging the vessels. Determination of
catheter interaction contact forces can improve the navigation process safety and efficiency, preventing injuries in both manual, and robotic vascular interventions \cite{razban2018sensor}. The force interaction of the catheter with the arterial wall can be estimated using an image-based algorithm combined with a force/deformation model of the catheter \cite{razban2020image}. The proposed PRB formulation can be used for the force/deformation model of the catheter in 2D. Additionally, as reported in \cite{razban2020image}, catheter has a variable stiffness along the length. We use our method for the analysis of such a problem in a separate case study.

In another scenario, we use the proposed method for the modeling of CTRs. The CTRs are comprised of super-elastic pre-curved tubes fitted inside each other in a telescopic way. 
The relative motion of the tubes (i.e., rotation and insertion) can be used to control the CTR shape and the position of its tip. 
CTRs are considered the smallest CR; hence they are a unique candidate for miniaturized surgeries such as retinal surgeries. CTRs in vitreoretinal surgery provide dexterity enhancement for controlling the tip angle of the instrument with respect to the retina. Additionally, they do not have the complication of the rigid robot, which imposes an extra force on the entry port at the sclera surface. Each section of the CTR can be considered as an HFM with a constant predefined initial curvature. CTRs generally are made of constant curvature segments due to the complexity of the variable curvature elements in both modeling and manufacturing. In what follows, we use the proposed method to obtain a PRB model for both constant curvatures HFM as well as variable curvature ones for 2D CTRs. The PRB model can be used for indirect force estimation of the CTR merely through the tip deflection.

Considering the above-mentioned case studies, in the remainder of this section, we provide the optimal values of stiffness for equal length PRB estimation over various ranges of DoF. However, for the first case study, we also consider the 3-DoF PRB models with non-identical segment lengths. In each case study, apart from the optimal parameters, the precision of the method is also analyzed by providing both position and force estimation errors. The errors are all in percentage and normalized, as follows:
\begin{equation}
\begin{array}{l}
\begin{cases}
\label{limit2}
e_{x}=\frac{1}{N}\sum_{q=1}^{N} \lvert{\frac{{x}_t-{\hat{x}_n}}{S}}\rvert^q\times 100\\\
e_{y}=\frac{1}{N}\sum_{q=1}^{N} \lvert{\frac{{y}_t-{\hat{y}_n}}{S}}\rvert^q\times 100\\\
e_{\theta}=\frac{1}{N}\sum_{q=1}^{N} \lvert\frac{{\theta}_t-{\hat{\theta}_{n+1}}}{{\theta^0_t}}\rvert^q\times 100
\end{cases}\\\
\begin{cases}
e_{f_x}=\frac{1}{N}\sum_{q=1}^{N} \lvert{\frac{{f}_x-{\hat{f}_x}}{max(f_x)}}\rvert^q\times 100\\\
e_{f_y}=\frac{1}{N}\sum_{q=1}^{N} \lvert{\frac{{f}_y-{\hat{f}_y}}{max(f_y)}}\rvert^q\times 100\\\
e_{m}=\frac{1}{N}\sum_{q=1}^{N} \lvert{\frac{{m_t}-{\hat m_t}}{max(m_t)}}\rvert^q\times 100
\end{cases}
\end{array}
\end{equation}
in which $[\hat{f_x},\hat{f_y},\hat{m_t}]^T=\bm{J}^{-T}\bm{K{\Delta\varphi}}$.

\subsection{PRB Modeling of Straight HFMs}
\label{straightHFMs}

\begin{figure}[ht]
	\centering
	\includegraphics[height=3.5cm,trim={6.5cm 12.3cm 7cm 12.3cm},clip=true]{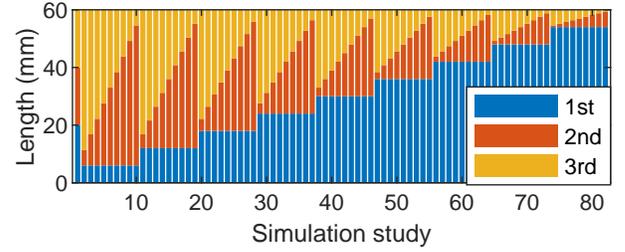}
	\caption{Various combinations of segment lengths for the 3-DoF PRB model}
	\label{combination of length}
\end{figure}

For the modeling of the catheter with the proposed approach, we consider a catheter with the parameters and loading conditions as described in \cite{venkiteswaran2019shape}. The maximum length of the catheter is chosen to be $50~mm$ mm with an elastic modulus $E$ of $350~MPa$ and an area moment of inertia $I=4.91 \times 10^{-2}~mm^4$. The maximum and minimum values of the tip force are $4~mN$ and $-4mN$, respectively, and the range of the tip moment is [$-250,-250$]~$mN.mm$. The analysis is done for $3410$ loading combinations within the above range. First, we investigate the performance of the proposed method for different segment lengths. This analysis is carried out for 82 combinations of segments $l_i$ such that $l_1+l_2+l_3=S$, as shown in Fig.~\ref{combination of length}.
  
\begin{figure}[ht]
	\centering
	\includegraphics[height=3.5cm,trim={6.5cm 12.3cm 7cm 12.3cm},clip=true]{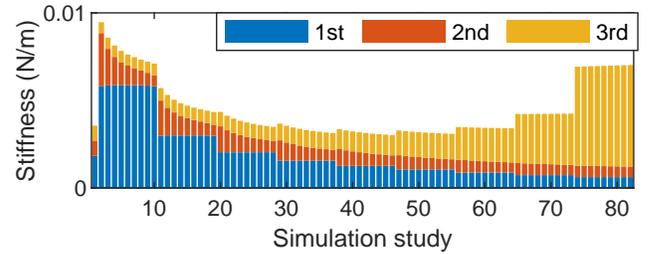}
	\caption{Stiffness values for the 3-DoF PRB models with various combinations of the segment lengths}
	\label{stiffness}
\end{figure}

For these simulations, the stiffness of each joint is summarized in Fig.~\ref{stiffness}. It can be observed from this figure that the stiffness of the joints increases by increasing the length of the corresponding segment, as expected physically. As for the HFM with uniform flexural rigidity along the length, the smaller the length of the segments, the larger the corresponding stiffness of the PRB model.

\begin{figure}[ht]
	\centering
	\includegraphics[height=3.4cm,trim={6.5cm 12.3cm 7cm 12.5cm},clip=true]{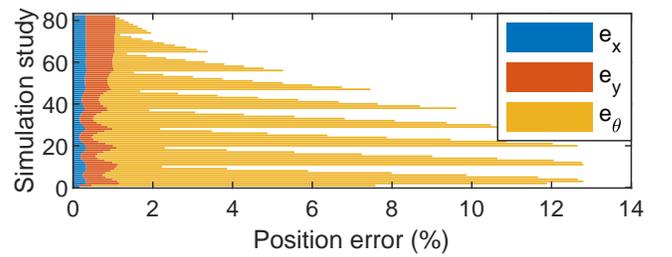}
	\caption{Position errors for the 3-DoF PRB models with various combinations of the segment lengths}
	\label{position error 3dof}
\end{figure}

\begin{figure}[ht]
	\centering
  	\includegraphics[height=3.6cm,trim={6.5cm 12.3cm 7cm 12.3cm},clip=true]{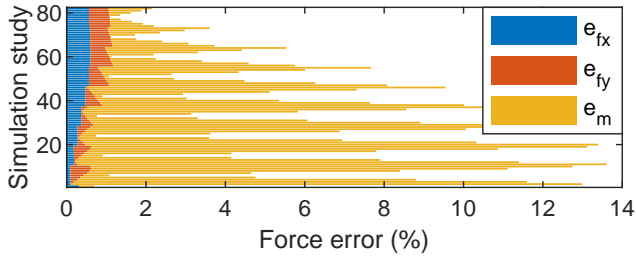}
	\caption{Force errors for the 3-DoF PRB models with various combinations of the segment lengths}
	\label{force error 3DoF}
\end{figure}

The position error of the PRB models for all the 82 configurations is summarized in Fig.~\ref{position error 3dof}. It is to be noted that for all the case studies, both the position and force errors are normalized and represented in percentage. As can be seen from the figure, in the first case (i.e., identical segment lengths), the total value of the error in Cartesian space (i.e., $e_x+e_y$) is minimum. On the other hand, the minimum error for the tip angle occurs when the length of the third segment is smaller, which is expected from the mechanical characteristics of the system. Distinguishing the position error as $e_x$, $e_y$, and $e_\theta$ provides an informed opportunity to choose the right PRB model for each application. 
The normalized force/moment percentage errors for all the combinations of the segment length are also depicted in Fig.~\ref{force error 3DoF}. 

Considering equal length for the segments, we have developed five different PRB models for the straight catheter. Each PRB model has a different DoF, and accordingly, the length of the segments are also different. The corresponding stiffness values for these PRB models are provided in Table~\ref{t1}. From these stiffness values, one can conclude that for each PRB model, there are only two values for the stiffness of the joints. The first value is for the stiffness of the first joint, and the second value is for that of the remaining joints (which have equal stiffness). It is to be noted that this result is only applied for the equal-length PRB models using the proposed methodology. This simplifies the analysis of the PRB model, especially when dealing with variable-length HFMs.

\begin{table}[ht]
	\caption{Optimal stiffness values for several PRB models of catheter}
\label{t1}
	\centering
\begin{tabular}{cccccc}\hline \hline
		DoF   & 3 & 4 & 10 & 15& 20\\ \hline
		$k_1$    & 0.0019& 0.0024& 0.0059& 0.0087& 0.0116 \\ \hline
		$k_2$   & 0.0009& 0.0011& 0.0029& 0.0043& 0.0057 \\ \hline
		$k_3$   & 0.0009& 0.0011 & 0.0029& 0.0043& 0.0057 \\ \hline
		$k_4$   & N/A & 0.0011& 0.0029& 0.0043& 0.0057 \\ \hline
		$k_{5-10}$    & N/A & N/A & 0.0029& 0.0043& 0.0057 \\ \hline
		$k_{15-20}$    & N/A & N/A & N/A & 0.0043& 0.0057 \\ \hline
	\end{tabular}
\end{table}

The force and position percentage error of each PRB model discussed above is also given in Table~\ref{t2}. These error values show that by increasing the DoF, all errors decrease. Therefore, one can choose the PRB model which best suits the specific application by adjusting the DoF properly.
\begin{table}[ht]

	\caption{Position and force errors of the catheter PRB models}
	\centering
\begin{tabular}{cccccc}\hline \hline
		DoF   & 3 & 4 & 10 & 15& 20\\ \hline
		$e_{px}$& 0.2081&0.1242&0.0189&0.0079&0.0041\\ \hline
		$e_{py}$&0.2956&0.1834&0.0266&0.0112&0.0058\\ \hline
		$e_{p\psi}$& 6.9509&5.3135&1.8691&1.08647&0.6990\\ \hline
		$e_{fx}$& 0.2683&0.2006&0.0772&0.0512&0.0382\\ \hline
		$e_{fy}$&0.0168&0.0129&0.0018&0.0008&0.0005\\ \hline
		$e_{m}$& 0.3738&0.2515&0.0410&0.0188&0.0106\\ \hline
	\end{tabular}
\label{t2}
\end{table}

 \subsection{PRB Modeling of Constant-curvature HFMs}  
 \label{section:fixed curvature}
In this subsection, we consider a case study using CTRs. An example of a miniaturized CTR is presented in \cite{lin2015biometry}, which is primarily developed for vitreoretinal surgeries. This CTR is comprised of two sections with the parameters specified in Table~\ref{param}. Considering nitinol as the material, the elastic modulus of the tubes is $71~GPa$.

\begin{table}[ht]
	\caption{Parameters of a CTR designed for eye surgery}
	\centering
\begin{tabular}{cccc}\hline \hline
		Tube type   & Inner tube & Outer tube\\ \hline
		Inner diameter (mm)    &0.203      &0.432 \\ \hline
		Outer diameter (mm)    &0.406      &0.635  \\ \hline
		Radius of curvature (mm)    &27      &80    \\ \hline
		Total length (mm)    &27         &26.5 \\ \hline
	\end{tabular}
	\label{param}
\end{table}

 For the case of membrane peeling, which is a retinal surgery, the range of forces is less than $7.5~mN$ \cite{gupta1999surgical}. Considering the force limit and the maximum length of $27$ mm for the curved element, the maximum torque for the PRB model is assumed to be $200~mN.mm$. The PRB model is to be optimized for the inner tube for 6820 lading cases over the range of loads ${f}\in[0,7.5]~mN$, $\psi\in[0,2\pi]$ and ${m_t\in[0,200]~mN.mm}$.

\begin{table}[ht]
	\caption{Optimal stiffness values of the CTR PRB models}
	\centering
\begin{tabular}{cccccc}\hline \hline
		DoF   & 3 & 4 & 10 & 15& 20\\ \hline
		$k_{1}$    &0.0205      &0.0271      &0.0666    &0.0995    &0.1323 \\ \hline
		$k_{2}$    &0.0099      &0.0132      &0.0329    &0.0493    &0.0658 \\ \hline
		$k_{3}$    &0.0099      &0.0132      &0.0329    &0.0493    &0.0658 \\ \hline
		$k_{4}$    &N/A         &0.0132      &0.0329    &0.0493    &0.0658 \\ \hline
		$k_{5-10}$    &N/A         &N/A         &0.0329    &0.0493    &0.0658 \\ \hline
		$k_{11-15}$   &N/A         &N/A         &N/A       &0.0493    &0.0658 \\ \hline
		$k_{16-20}$   &N/A         &N/A         &N/A       &N/A       &0.0658 \\ \hline
	
	\end{tabular}
	\label{t4}
\end{table}
As it can be observed from the results in Table~\ref{t4}, for this case study, similar to first one, which used the catheter PRB model, there are only two values for the stiffness of the joints, regardless of the DoF. In Table~\ref{t5} the values of the tip position error and force estimation error are for this PRB model.

\begin{table}[ht]
	\caption{Position and force errors of CTR PRB models}
	\centering
\begin{tabular}{cccccc}\hline \hline
		DoF   & 3 & 4 & 10 & 15& 20\\ \hline
		$e_{px}$   &  0.3897  &  0.2183  &  0.0337  &  0.0142 &   0.0074 \\ \hline
		$e_{py}$   &  0.2137 &   0.1197  &  0.0185  &  0.0078 &   0.0040
		\\ \hline
		$e_{p\theta}$& 0.5353  &  0.3919 &   0.1357  &  0.0791 &   0.0508\\ \hline
		$e_{fx}$ &0.1373  &  0.1019  &  0.0400 &   0.0265 &   0.0199\\ \hline
		$e_{fy}$ &0.0168 &   0.0096  &  0.0016  &  0.0007  &  0.0004\\ \hline
		$e_{m}$  &0.5948 &   0.3428  &  0.0569 &   0.0255  &  0.0144\\ \hline
		
	\end{tabular}
	\label{t5}
\end{table}

The results show that by increasing DoF of the PRB model, the overall error for both position and force decreases. 


\subsection{PRB Modeling of Variable-curvature HFMs}
As discussed in Section~\ref{PRB}, one of the limitations of the available PRB modeling methods is in the modeling of HFMs, where the curvature changes along the length. To the best of author's knowledge, there is no PRB research that can directly obtain PRB model for HFM with variable curvature along the length. On the other hand, for the HFM with a circular shape, while it is theoretically assumed that the curvature is constant along the length, irregularities in the shape are inevitable during the manufacturing process. Additionally, the mechanical elements will have plastic deformation over time which appears as a change of shape. With a variable curvature HFM, there is more flexibility in the optimal design of HFM-based systems, such as constant force grippers. However, the limitations of the conventional methods complicate the possibility of the parametric design of nonuniform curvature HFMs.
In the sequel, we investigate the performance of the proposed PRB modeling approach for the HFMs with a variable curvature along the length. As an example, we consider the curvature to be linearly varying along the length, i.e., $r(s)= a\cdot s + b$, where $a$ and $b$ are known constants. Let the curvature be $27~mm$ at the base and $9~mm$ at the tip point. The loading condition and all the other parameters are similar to the previously presented fixed curvature case study for the CTR. The optimal values of the stiffness of the PRB model are given in Table~\ref{t6}, which show that except for the 3-DoF model, all other ones have two values for the stiffness of the PRB model.  
\begin{table}[ht]
	\caption{Optimal stiffness values for PRB models of a HFM with variable curvature along the length}
	\centering
\begin{tabular}{cccccc}\hline \hline
		DoF   & 3 & 4 & 10 & 15& 20\\ \hline
		$k_{1}$    &0.0205      &0.0270      &0.0665    &0.0993    &0.1319 \\ \hline
		$k_{2}$    &0.0099      &0.0132      &0.0329    &0.0493    &0.0658 \\ \hline
		$k_{3}$    &0.0130      &0.0132      &0.0329    &0.0493    &0.0658 \\ \hline
		$k_{4}$    &N/A         &0.0132      &0.0329    &0.0493    &0.0658 \\ \hline
		$k_{5-10}$    &N/A         &N/A         &0.0329    &0.0493    &0.0658 \\ \hline
		$k_{11-15}$   &N/A         &N/A         &N/A       &0.0493    &0.0658 \\ \hline
		$k_{16-20}$   &N/A         &N/A         &N/A       &N/A       &0.0658 \\ \hline
	\end{tabular}
	\label{t6}
\end{table}
In Table~\ref{t7}, the values of the tip position error and force estimation error have been represented for the PRB model.

\begin{table}[ht]
	\caption{Position and force errors for PRB models of a HFM with variable curvature along the length}

	\centering
\begin{tabular}{cccccc}\hline \hline
		DoF   & 3 & 4 & 10 & 15& 20\\ \hline
$e_{px}$   &  5.9569  &  0.3590 &   0.0584  &  0.0247 &0.0129 \\ \hline
$e_{py}$   &  0.9082  &  1.1030  &  0.1697  &  0.0714  &  0.0371
   \\ \hline
$e_{p\theta}$& 0.8184  &  0.3919  &  0.1357 &   0.0791  &  0.0508\\ \hline
$e_{fx}$ &0.1868  &  0.1001   & 0.0403  &  0.0269 &   0.0202\\ \hline
$e_{fy}$ &0.0420 &   0.0199 &   0.0030  &  0.0013 &   0.0007\\ \hline
$e_{m}$  &2.0934  &  0.7897  &  0.1158  &  0.0505  &  0.0281\\ \hline

	\end{tabular}
	\label{t7}
\end{table}

\subsection{PRB Modeling of Straight HFMs with Nonuniform Stiffness}  
 The flexural stiffness is represented by the product of $E$ and $I$. Thus, if the cross-section of the HFMs is nonuniform or if the material properties are not uniform along the length, the flexural stiffness would be variable. This is the case for the catheter, where the flexural stiffness decreases from the base to the tip point. As a case study, we consider the catheter described in \ref{straightHFMs}
 under a similar loading condition and let the stiffness be variable along the length. For this analysis, the distribution of the stiffness is assumed to be linearly varying along the length, expressed as $EI(s)=-\frac{E\cdot I}{2S}s+E\cdot I$.
 
The results of the PRB model for various DoF is presented in Table~\ref{t9}. As expected we have different values for the joint stiffness in each PRB model.
\begin{table}[ht]
	\caption{Optimal stiffness values of the PRB models for a HFM with non-uniform stiffness}
	\centering
\begin{tabular}{cccccc}\hline \hline
		DoF   & 3 & 4 & 10 & 15& 20\\ \hline
		$k_{1}$    &0.0063      &0.0086      &0.0224    &0.0339    &0.0454 \\ \hline
		$k_{2}$    &0.0028      &0.0040      &0.0109    &0.0166    &0.0224 \\ \hline
		$k_{3}$    &0.0022      &0.0034      &0.0103    &0.0160    &0.0218 \\ \hline
		$k_{4}$    &N/A         &0.0028      &0.0097    &0.0155    &0.0212 \\ \hline
		$k_{5}$    &N/A         &N/A         &0.0092    &0.0149    &0.0206 \\ \hline
		$k_{6}$    &N/A         &N/A         &0.0086    &0.0143    &0.0201 \\ \hline
		$k_{7}$    &N/A         &N/A         &0.0080    &0.0137    &0.0195 \\ \hline
		$k_{8}$    &N/A         &N/A         &0.0074    &0.0132    &0.0189 \\ \hline
		$k_{9}$    &N/A         &N/A         &0.0069    &0.0126    &0.0183 \\ \hline
		$k_{10}$   &N/A         &N/A         &0.0063    &0.0120    &0.0178 \\ \hline
		$k_{11}$   &N/A         &N/A         &N/A       &0.0114    &0.0172 \\ \hline
		$k_{12}$   &N/A         &N/A         &N/A       &0.0109    &0.0166 \\ \hline
		$k_{13}$   &N/A         &N/A         &N/A       &0.0103    &0.0160 \\ \hline
		$k_{14}$   &N/A         &N/A         &N/A       &0.0097    &0.0155 \\ \hline
		$k_{15}$   &N/A         &N/A         &N/A       &0.0092    &0.0149 \\ \hline
		$k_{16}$   &N/A         &N/A         &N/A       &N/A       &0.0143 \\ \hline
		$k_{17}$   &N/A         &N/A         &N/A       &N/A       &0.0137 \\ \hline
		$k_{18}$   &N/A         &N/A         &N/A       &N/A       &0.0132 \\ \hline
		$k_{19}$   &N/A         &N/A         &N/A       &N/A       &0.0126 \\ \hline
		$k_{20}$   &N/A         &N/A         &N/A       &N/A       &0.0120 \\ \hline
	\end{tabular}
	\label{t9}
\end{table}
The error of PRB approximation for this case study is provided in Table~\ref{t8}. 

\begin{table}[ht]
	\caption{Position and force errors for the PRB models of a HFM with non-uniform stiffness}
\centering
\begin{tabular}{cccccc}\hline \hline
		DoF   & 3 & 4 & 10 & 15& 20\\ \hline
	$e_{px}$    &0.0051    &0.0028    &0.0004    &0.0002    &0.0001 \\ \hline
	$e_{py}$   &0.1112    &0.0633    &0.0099    &0.0042    &0.0022\\ \hline
	$e_{p\theta}$    &3.1931    &2.4242    &0.8994    &0.5328    &0.3452\\ \hline
	$e_{fx}$    &0.0827    &0.0635    &0.0317    &0.0261    &0.0238\\ \hline
	$e_{fy}$    &0.0144    &0.0139    &0.0169    &0.0178    &0.0183\\ \hline
	$e_{m}$    &0.3195    &0.2203    &0.3410    &0.3847    &0.4071\\ \hline	
\end{tabular}
\label{t8}
\end{table}

\subsection{PRB Modeling of the Initially-curved Variable-length HFMs}
\label{variablelengthfixedcurvature}
\begin{figure}[ht]
	\centering
	\includegraphics[height=7cm,trim={4cm 8.3cm 4.5cm 9.1cm},clip=true]{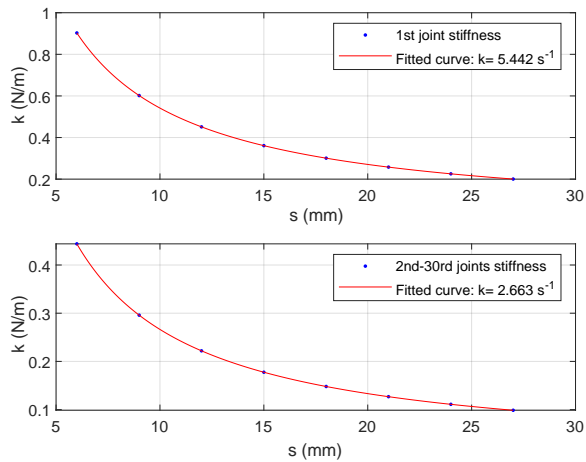}
	\caption{Optimal stiffness for 30-DoF PRB models of a variable-length HFM as a function of HFM length}
	\label{Stiffnessvariablelength}
\end{figure}

In CTRs, the initially curved segments are telescopically constrained, and the relative position of the segments allows the tip of the robot to follow a specified trajectory. Thus, a suitable model of the robot should take into account the variable length of each curved segment. From the results obtained in the previous case studies, the PRB model for the equal length of the segments is specified with three parameters. These three parameters are $k_1$, which is the stiffness of the first joint, $k_2$, which is the stiffness of the other joints, and the length of each segment (i.e., $S/n$). Considering this fact, $k_1$ and $k_2$ can be found as functions of the total length of the curved segment while keeping the DoF of the PRB model the same. In other words, for a HFM with the length of $S=L$ first we develop an $n$-Dof PRB model with ${k_1}$, ${k_2}$ and $l=L/n$ as the PRB model parameters, then we develop another $n$-DoF PRB model for the same HFM which has a different length , i.e. $S = L + \Delta L$, with ${k_1}^\prime$, ${k_2}^\prime$ and $l^\prime=(L + \Delta L)/n$ being the parameters of the PRB model. By repeating this procedure for several $\Delta L$, one can find a function that gives $k_1$ and $k_2$ as a function of HFM length. Fig.~\ref{Stiffnessvariablelength} shows the stiffness values of a 30-DoF PRB model as a function of the HFM length. This study shows that the stiffness values of the PRB model  for variable-length HFM (i.e., $k_1$ and $k_2$), with the proposed approach, can be represented with a power function, represented as follows:
\begin{equation}
k=\kappa s^{-\sigma}
\label{stiffnessformula}
\end{equation}
where $\kappa$ and $\sigma$ are the tuning variables.

\section{Conclusion}
\label{conclusion}
In this paper, we proposed a semi-analytical method in literature for the PRB modeling of the highly flexible members (HFM). The proposed method overcomes the limitations of the conventional methods for variable-curvature HFM. Additionally, it is very simple to use for the researchers and has the minimum number of parameters for the PRB model, regardless of the DoF. Furthermore, the errors of the model for each component of the force and position can be easily studied with the proposed method. Applying the proposed method for the PRB modeling of the HFM shows the versatility and comprehensiveness of the method with a better precision compared to the available methods in the literature. It is notable that PRB methods are generally developed for 2D problems, which is the same as the current study. However, in robotic, we required models to be used for 3D elements as well, which can be considered as the extension of the proposed method. For future work, we will use the proposed method for the interaction force estimation of the HFMs with the environment and will extend the model for spatial HFMs.


\bibliographystyle{unsrt}

\end{document}